\pdfoutput=1
\documentclass[10pt,twocolumn,letterpaper]{article}

\usepackage{multirow}
\usepackage{amsmath}
\usepackage{array}
\usepackage[export]{adjustbox}
\usepackage[pagenumbers]{cvpr}

\usepackage[symbol]{footmisc}

\usepackage[dvipsnames]{xcolor}

\definecolor{cvprblue}{rgb}{0.21,0.49,0.74}
\usepackage[pagebackref,breaklinks,colorlinks,citecolor=cvprblue]{hyperref}

\def\paperID{4536} 
\def\confName{CVPR}
\def\confYear{2024}

\title{Robust Depth Enhancement via Polarization Prompt Fusion Tuning}

\author{Kei Ikemura$^{6}$\thanks{Equal contribution} \quad Yiming Huang$^{5}$\footnotemark[1] \quad Felix Heide$^{2}$ \quad \\Zhaoxiang Zhang$^{1,4}$  \quad Qifeng Chen$^3$ \quad Chenyang Lei$^{1,2}$\thanks{Corresponding author} \vspace{0.3em} \\
{\small{ $^1$CAIR, HKISI-CAS} \quad
{ $^2$Princeton University} \quad 
{ $^3$HKUST} \quad} 
{\small{ $^4$CASIA}  \quad 
{ $^5$CUHK} \quad 
{ $^6$KTH Royal Institute of Technology} \quad} 
}

\begin{document}

\maketitle

\begin{abstract}
Existing depth sensors are imperfect and may provide inaccurate depth values in challenging scenarios, such as in the presence of transparent or reflective objects. In this work, we present a general framework that leverages polarization imaging to improve inaccurate depth measurements from various depth sensors. Previous polarization-based depth enhancement methods focus on utilizing pure physics-based formulas for a single sensor. In contrast, our method first adopts a learning-based strategy where a neural network is trained to estimate a dense and complete depth map from polarization data and a sensor depth map from different sensors. To further improve the performance, we propose a Polarization Prompt Fusion Tuning (PPFT) strategy to effectively utilize RGB-based models pre-trained on large-scale datasets, as the size of the polarization dataset is limited to train a strong model from scratch. We conducted extensive experiments on a public dataset, and the results demonstrate that the proposed method performs favorably compared to existing depth enhancement baselines. Code and demos are available at \url{https://lastbasket.github.io/PPFT/}.

\end{abstract}

\section{Introduction}
\label{sec:intro}
\setlength{\belowcaptionskip}{-15pt}
\begin{figure}[ht!]

	\centering
 \includegraphics[width=0.85\linewidth]{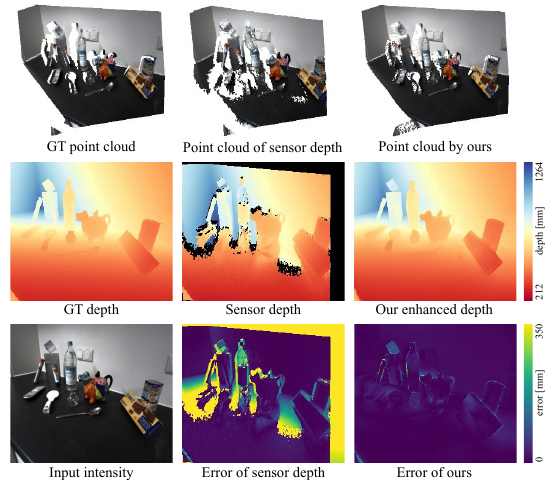}
  \vspace{-2mm}

 \caption{\textbf{Visualization of the results of our proposed framework}. The sensor depth shown is from a d-ToF sensor. Our method leverages the dense shape cues from polarization and produces accurate results on challenging depth enhancement problems, including depth in-painting, restoring depth on transparent surfaces, shape correction, etc. For example, see that the depth of the transparent water bottle is restored. Note that on the error map of the sensor depth, we clamp values to the maximum scale used.} 
\label{fig:Examples}
\end{figure}
\setlength{\belowcaptionskip}{+15pt}

Estimating accurate depth is crucial for a wide range of applications, from autonomous driving to 3D reconstruction. Over the past decades, various depth sensing methods have been investigated, including Direct \cite{qiu2019deeplidar, xu2019depth} and Indirect \cite{jung2021wild, horio2022resolving, schelling2023weakly} Time-of-Flight (ToF) cameras, structured light sensors \cite{gupta2011structured, gupta2013structured}, and stereo cameras \cite{garcia2012depth}. Despite wide application, these sensing systems live in a complex trade-off space where each has its own vulnerabilities, including specularities \cite{or2016real}, transparency \cite{chai2020deep}, Multi-Path-Interference (MPI)~\cite{horio2022resolving, schelling2023weakly}, strong ambient light~\cite{gupta2011structured, gupta2013structured}, and texturelessness \cite{chai2020deep} just to name a few.

In this work, we are interested in using polarization data as generalizable geometry cues for depth enhancement. The polarization state is closely related to the local surface orientation, providing valuable information for accurate depth estimation. Moreover, as a passive sensor, polarization cameras have the advantage of capturing dense geometry cues without being limited by the depth range \cite{lei2022shape}. Further, polarization measurement can be used to detect abnormal conditions such as reflection \cite{lei2022shape, mei2022glass} and transparency \cite{kalra2020deep}. This can be significantly reliable for detecting and correcting false depth measurements. Hence, polarization can serve as a highly effective complementary component to existing depth sensors, enhancing their overall performance and expanding their capabilities in various applications.

To the best of our knowledge, no existing work focuses on a general depth enhancement that applies to various depth-sensing technologies using polarization guidance. Some previous studies have explored physics-based methods~\cite{kadambi2015polarized, yoshida2018improving}, achieving accurate depth enhancement results. However, they are limited by several strict assumptions. For example, Yoshida \textsl{et al.} \cite{yoshida2018improving} assumed specular
reflection and that the target object in interest represents a closed surface. Other studies leveraged deep learning with polarization to address particular depth sensing problems, including texturelessness for stereo-vision \cite{Tian_2023_ICCV}, and multi-path inference for i-ToF sensors \cite{jeon2023polarimetric}. Despite the robust enhancement result, they are limited to specific depth sensing technology. In this work, we aim to extend the capability of deep learning techniques to achieve robust polarization-guided depth enhancement that generalizes across different sensing modalities.
However, we noticed that learning with polarization for general depth enhancement is not trivial. From our primary experiments, we observe little performance gains by directly passing polarization into existing RGB-based depth enhancement models, as illustrated in Table \ref{table:non-trivial}. Moreover, the size of existing polarization datasets \cite{hammer, verdie2022cromo},  are not comparable to the large-scale RGB-D datasets \cite{Silberman-ECCV12, uhrig2017sparsity}, resulting in a limited size of the training data available for polarization. On this, the naive approach of collecting as much polarization data as RGB images alleviates the problem but is unfavorable as it is extremely labor-intensive. 

\begin{table}[!t]
\footnotesize
\centering
\renewcommand{\arraystretch}{1.2}
\setlength{\belowcaptionskip}{-13pt}
\resizebox{0.85\linewidth}{!}{
\begin{tabular}{p{2.7cm}   | p{0.6cm} p{0.6cm} p{0.6cm} p{0.6cm} p{0.6cm}}
\toprule[1pt]
 Model &  RMSE$\downarrow$ & MAE$\downarrow$  & $\delta_1 \uparrow$ &$\delta_2 \uparrow $ & $\delta_3 \uparrow $\\
  Without polarization & 43.64 & 22.84 & 0.978 & 0.999 & 0.999\\

  Naive with polarization & 41.17 & 22.14 & 0.981 &	0.999 & 0.999 \\
  
  Ours with polarization  &\textbf{28.63} & \textbf{14.17}	& \textbf{0.992} & \textbf{0.999} & \textbf{1.000} \\
\toprule[1pt]
\end{tabular}}
  \vspace{-2mm}
\caption{\textbf{Naively incorporated polarization on HAMMER~\cite{hammer} Dataset.} Results are average on all sensors and "Without polarization" indicates using the baseline~\cite{youmin2023completionformer}. One observes little performance gains by naively passing polarization data into RGB-guided methods directly. In contrast, our approach could effectively utilize the polarization, achieving significantly better performance.}

\label{table:non-trivial}
\end{table}
\setlength{\belowcaptionskip}{+13pt}

To overcome these issues, we propose a cross-modal transfer learning strategy, Polarization Prompt Fusion Tuning (PPFT). First, to prevent overfitting on small polarization datasets, we incorporate pre-trained model weights from large-scale RGB-D datasets into our backbone. Second, we target the modality misalignment by the proposed Polarization Prompt Fusion Block (PPFB), a parallel branch design where a spatial attention-like operation and a channel fovea operation are introduced. Utilizing our PPFB throughout the feature extraction network, the polarization cues can be integrated into the network more effectively.

\begin{figure}[t]
\centering
\begin{tabular}{@{}c@{\hspace{1.1mm}}c@{\hspace{1.1mm}}c@{\hspace{1.1mm}}c@{\hspace{1.1mm}}c@{\hspace{1.1mm}}c@{\hspace{1.1mm}}c@{\hspace{1.1mm}}c@{}}
\includegraphics[width=0.23\linewidth]{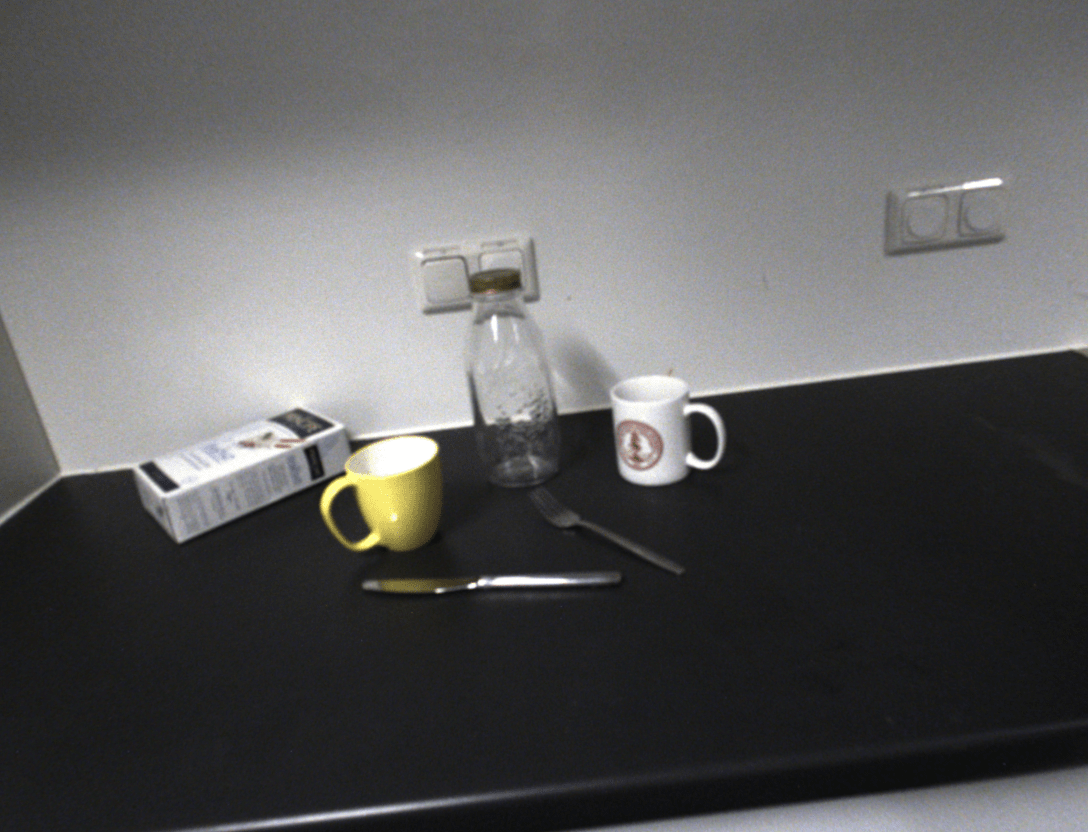}&
\includegraphics[width=0.23\linewidth]{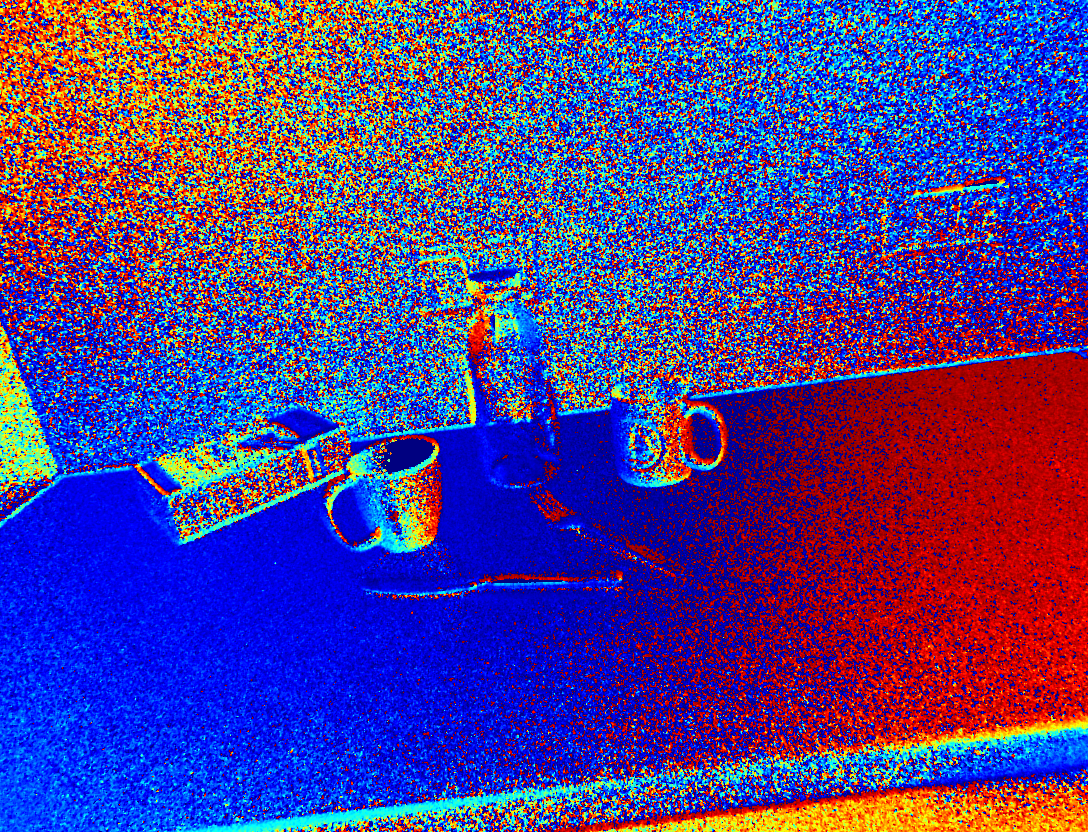}&
\includegraphics[width=0.23\linewidth]{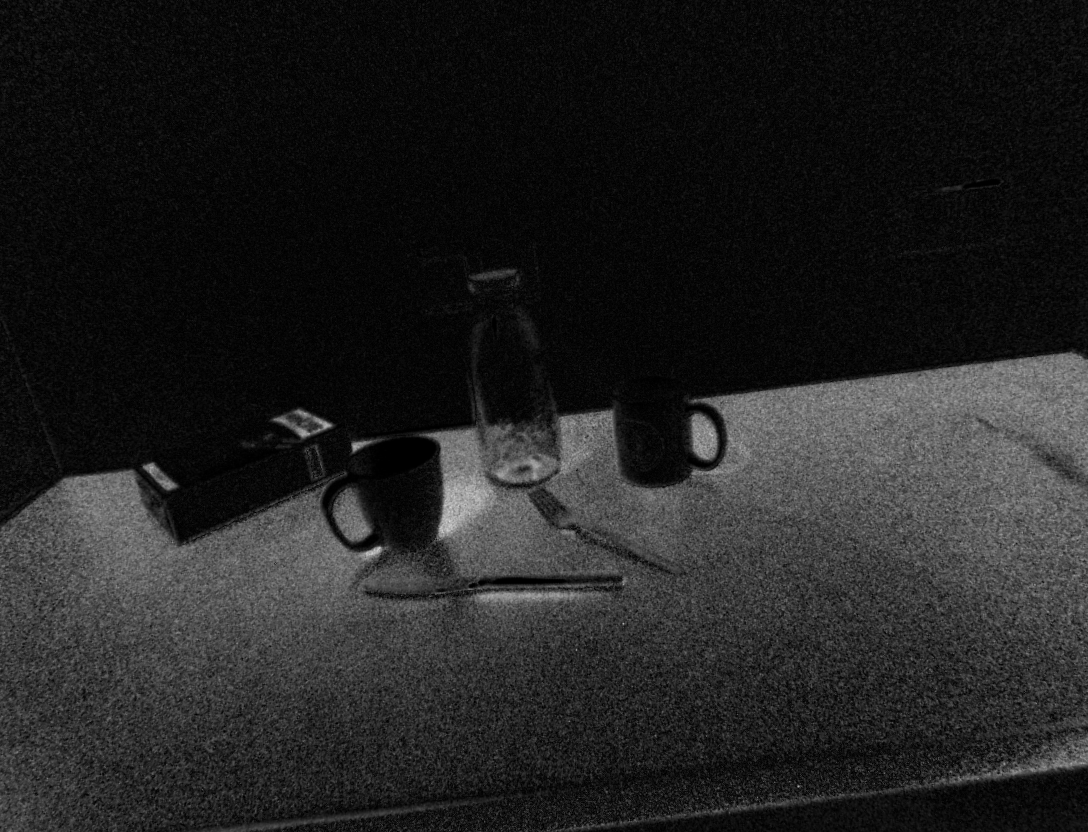}&
\includegraphics[width=0.23\linewidth]{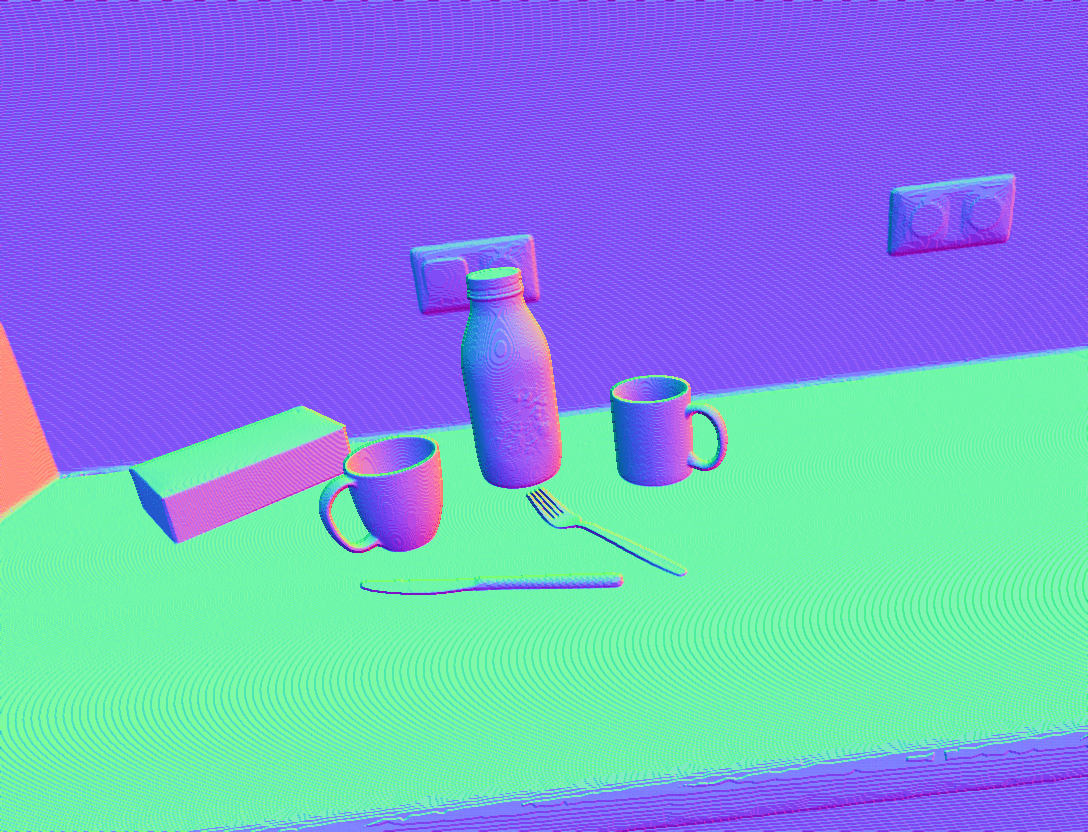}

\\
  \scriptsize{(a) RGB} &  \scriptsize{(b) AoLP}&  \scriptsize{(c) DoLP} &  \scriptsize{(d) GT Normal}  \\ 
 \includegraphics[width=0.23\linewidth]{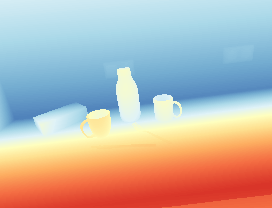}&
\includegraphics[width=0.23\linewidth]{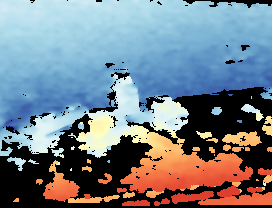}&
\includegraphics[width=0.23\linewidth]{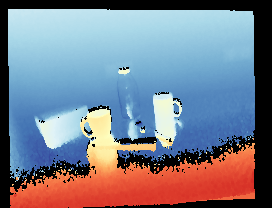}&
\includegraphics[width=0.23\linewidth]{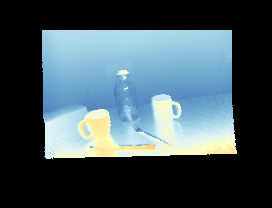}\\
 \scriptsize{(e) GT Depth} &  \scriptsize{(f) Stereo} &  \scriptsize{(g) d-ToF} &  \scriptsize{(h) i-ToF} \\
\end{tabular}
\\

\vspace{-2mm}
  \caption{\textbf{Visualization of our evaluation data by \cite{hammer}}. One can observe varying types of data degradation in each depth sensor type. For example, in (f) we can see missing measurements at the table surface, due to insufficient texture. Similarly, as the dataset provides all sensor depth mapped to the same camera frame, in (h) we observe a region of unknown depth at the boundary. This arises from the smaller Field-of-View (FoV) of the i-ToF sensor. In this work, we propose a general framework to resolve multiple types of sensor depth degradation.}
\label{fig:low-quality-patterns}
\vspace{-5mm}
\end{figure}

We conduct extensive experiments on different depth sensor types to validate the effectiveness of our approach. Examples of various depth modalities are shown in Figure \ref{fig:low-quality-patterns}. We find that our model significantly improves the quality across all types of sensor depths with the help of polarization. While other methods fail in challenging cases such as in the presence of transparent objects, reflective surfaces, etc., our enhanced depth maps are more robust and can restore detailed geometry.

In summary, we make the following contributions
\begin{itemize}
    \item We devise a general depth enhancement approach that leverages polarization information as dense shape cues. To the best of our knowledge, our method is the first polarization-based depth enhancement method that can handle different types of sensor depth.

    \item To tackle data scarcity in existing polarization datasets, we propose a novel cross-modal transfer learning strategy named Polarization Prompt Fusion Tuning (PPFT) for polarization-based vision. PPFT improves the performance significantly on our depth enhancement task. Besides, we also prove its effectiveness on another polarization-based task, Shape-from-Polarization.

    \item We conduct extensive experiments that validate our method outperforms state-of-the-art methods on different types of depth sensors. We provide ablation studies to understand the effect of polarization and various components of our proposed method.
\end{itemize}

\setlength{\belowcaptionskip}{-1pt}
\section{Related Work}
\label{sec:related-works}

\paragraph{Depth Enhancement via Polarization Cues}
Studies~\cite{cui2017polarimetric, berger2017depth, zhu2019depth} have demonstrated the usefulness of polarization in assisting multi-view stereo depth, particularly in textureless regions. Kadambi \textsl{et al}.~\cite{kadambi2017siggraph, kadambi2015polarized, kadambi2017depth} enhanced coarse depth map from structure light cameras by using the shape information from polarization cues. On the other hand, \cite{reda2017polarization, dashpute2018depth, yoshida2018improving, zhang2022time} employed polarization to refine incomplete depths from Time-of-Flight sensors, and Jeon \textsl{et al.} \cite{jeon2023polarimetric} proposed to address incorrect depth measurement given scattering media by polarization data. In addition, Shakeri \textsl{et al.}~\cite{shakeri2021polarimetric} explored the dense depth map reconstruction with polarization. Recently, \cite{Tian_2023_ICCV} extended the polarimetric stereo depth using deep learning method.  In contrast, we propose a polarization-based deep learning method for depth enhancement generalized to multi-sensor depths.

\vspace{-6pt}
\paragraph{Polarization-Guided Vision Task}
 We can also draw insight from works that study alternative vision tasks with polarization. \cite{kalra2020deep, mei2022glass, Liang_2022_CVPR} utilized polarization data for object segmentation, fusing the angle of linear polarization, degree of linear polarization, and intensity image to achieve material-specific segmentation, such as on transparent and semi-transparent materials. \cite{Ichikawa_2021_CVPR, chen2022perspective} focus on physics-based polarimetric 3D methods, while \cite{ba2020deep, lei2022shape, Muglikar_2023_CVPR, Huang_2023_CVPR, Shao_2023_ICCV, deschaintre2021deep} explored object-level and scene-level Shape-from-Polarization (SfP) with deep learning. In addition, \cite{kondo2020accurate, dave2022pandora} demonstrate the utilization of polarization in rendering. On low-level vision tasks, \cite{wieschollek2018separating, lei2020polarized} leveraged polarization properties of light and deep learning for specular reflection removal. Ono \textsl{et al.}~\cite{Ono_2022_CVPR} proposed to solve color constancy with DoLP, and \cite{Li_2023_CVPR, Kurita_2023_WACV} contributed to novel polarization enhancement approaches. By drawing insights from previous polarimetric vision tasks, we focus on the fusion and complement of polarization and coarse depth information, mitigating ambiguity with deep learning.

\vspace{-6pt}
\paragraph{RGB-Based Depth Enhancement} \label{depth-comp}
A major line of research for depth enhancement investigates deep learning with (RGB and/or intensity) images as guidance~\cite{zhang2018deep, park2020non, rho2022guideformer, youmin2023completionformer, wang2023lrru, liu2021fcfr, qiu2019deeplidar, xu2020deformable, xu2019depth, nazir2022semattnet}. Some works have studied a coarse-to-fine strategy with spatial propagation networks (SPN)~\cite{liu2017learning}, by learning spatial affinities~\cite{cheng2018depth, park2020non, cheng2020cspn++, xu2020deformable}. Further, geometric constraints are also exploited in the depth enhancement process. Zhang \textsl{et al}.~\cite{zhang2018deep} used a deep CNN to predict the occlusion boundary and surface normal to regularize the depth enhancement process. Qiu \textsl{et al}. \cite{qiu2019deeplidar} and Xu \textsl{et al}. \cite{xu2019depth} proposed end-to-end deep learning frameworks with surface normal as intermediate representations to guide depth enhancement. Following recent advancements in Transformer architectures~\cite{vaswani2017attention,dosovitskiy2020image}, Rho \textsl{et al}. proposed GuideFormer~\cite{rho2022guideformer}, a Transformer-based depth completion model with RGB image guidance. More recently, Zhang \textsl{et al}. combined convolution operation and Transformer layer in a unified block \cite{youmin2023completionformer} for depth completion.

Although there have been numerous studies on depth enhancement using image guidance and deep learning methods, alternative modalities-guided processes have received little attention. In this paper, we propose to use polarization as guidance, offering superior surface geometry information and enabling the detection of irregular conditions.

\raggedbottom

\setlength{\belowcaptionskip}{+1pt}

\section{Method}

\setlength{\belowcaptionskip}{-4pt}
\begin{figure*}[!ht]
  \centering
  \includegraphics[width=0.9\linewidth]{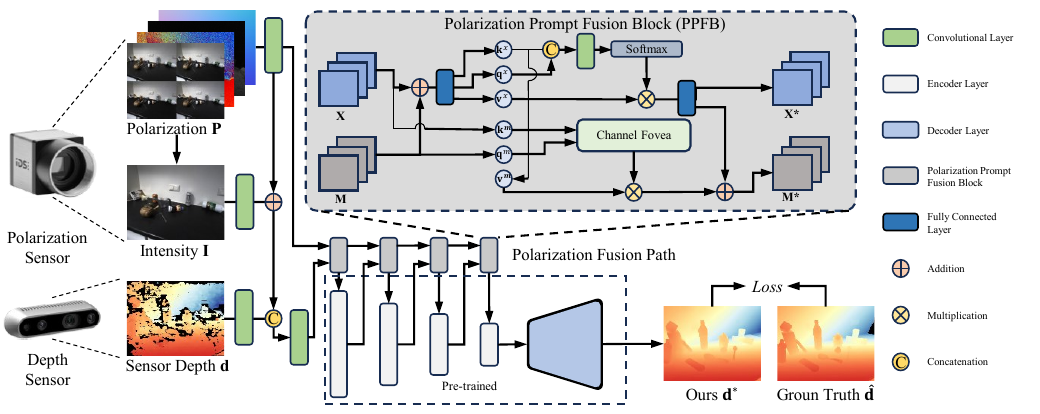}
    \vspace{-2mm}
  \caption{\textbf{Polarization Prompt Fusion Tuning (PPFT).} We fuse polarization embeddings to the features extracted from pre-trained layers sequentially using our proposed Polarization Prompt Fusion Block (PPFB). Specifically, polarization features are passed into our PPFB as prompt $\mathbf{M}$, and features from the pre-trained foundation as the input $\mathbf{X}$. Both are then updated and passed into the next set of pre-trained encoder and our PPFB respectively.}
  \label{fig:framework}
\end{figure*}
\setlength{\belowcaptionskip}{+4pt}
\setlength{\belowcaptionskip}{-11pt}
\begin{figure}[!t]
    \centering
    \includegraphics[width=0.85\linewidth]{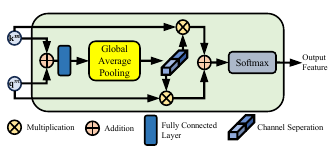}
      \vspace{-2mm}
    \caption{\textbf{Channel Fovea Operation in Proposed PPFB.} The summation of the input features is transformed to have doubled channel size with an MLP, and then distilled to a list of weights by global average pooling \cite{lin2013network}. These weights are separated into two sets and multiplied with the two inputs, respectively. Output attention weights are computed via a softmax operation on the sum of the re-weighted inputs.}
    \label{fig:channel-fovea}
\end{figure}

\setlength{\belowcaptionskip}{+11pt}

\label{sec:method}
We propose the first end-to-end learning-based general depth enhancement method with polarization guidance. Figure \ref{fig:framework} illustrates the proposed method. The method operates on the raw sensor depth $\mathbf{d}$, as well as the polarization data $\mathbf{P}$. In the following, we first present the preliminaries of polarization imaging as it relates to depth enhancement in Section~\ref{sec:polar}. We then describe the proposed learning-based polarization-guided depth enhancement method in Section~\ref{sec:arch}. Finally, we describe our polarization prompt fusion tuning (PPFT) strategy in Section~\ref{sec:finetuning}.

\subsection{Polarization as General Depth Guidance}
\label{sec:polar}
A division-of-focal-plane (DoFP) polarization camera measures the light intensity $\mathbf{I}_{\mathrm{pol}}$ passing through the linear polarizer at polarization angle $\phi_{\mathrm{pol}}$ is given by~\cite{verdie2022cromo, lei2022shape, ba2020deep}:
\begin{equation}
\label{eq:I_pol}
    \mathbf{I}_{\mathrm{pol}} = \mathbf{I}_{\mathrm{un}} * (1 + \rho * cos(2\phi - 2\phi_{\mathrm{pol}})),
\end{equation}
where $\mathbf{I}_{\mathrm{un}}$ is the total incident light into the camera, which is unpolarized in general, $ \rho$ is the degree of linear polarization (DoLP), and $\phi$ is the angle of linear polarization (AoLP). 

Polarization is affected by the surface geometry of scenes, which can, in turn, provide crucial guidance to depth enhancement. Specifically, assuming a perspective projection, the AoLP $\mathbf{\phi}$, and corresponding surface normal $\mathbf{n}$, can be related via \cite{lei2022shape}:$$
\begin{cases}
        \mathbf{\Phi} = \mathbf{n} \times \mathbf{v} \times [0,0,1]^T, & \text{ if diffuse dominant}\\
        
        \mathbf{\Phi} = \mathbf{n} \times \mathbf{v} \times \mathbf{v} \times [0,0,1]^T, & \text{ if specular dominant}
    \end{cases}
$$ and $$\phi = \text{arctan}(\frac{\Phi_y}{\Phi_x}),$$ where $\Phi = [\Phi_x, \Phi_y, \Phi_z]^T$, and $\mathbf{v} \in 	\mathbb{R}^3$ is the viewing direction. Though there is a strong relationship between AoLP and surface normal, directly deriving the surface normal vector given polarization data is challenging under the perspective projection assumption \cite{lei2022shape}.

In this work, we rely on the strong correlation between surface normal and polarization data and investigate a learning-based strategy to extract and exploit such guidance for depth enhancement. The polarization guidance is obtained by concatenating the intensity image $\mathbf{I}$, the angle of polarization image $\phi$, the degree of polarization image $\rho$, and the viewing direction $\mathbf{V}$ as
\begin{equation}
        \mathbf{P} = [\mathbf{I} ; \phi ;  \rho ; \mathbf{V}],
\end{equation}
where $[\cdot]$ denote concatenation. A typical DoFP polarization camera captures a raw image $\mathbf{I}$ consisting of four polarization images under different polarizer angles: $0, \frac{\pi}{4}, \frac{\pi}{2}, \frac{3\pi}{4}$. One can recover the complete linear polarization state $\phi$ and $\rho$ following Equation~\ref{eq:I_pol}. As most depth sensors capture scene-level data, the commonly adopted orthographic projection assumption \cite{wolff1990surface, rahmann2001reconstruction, ba2020deep} fails for the setting targeted in this work and we assume a perspective projection. Since under such an assumption, the polarization state is closely related to the viewing direction $\mathbf{V}$, we pass it along with the polarization states. Further details are discussed in the Supplementary Material. 

\subsection{Learning Polarization Depth Enhancement}
 \label{sec:arch}

Polarization provides strong cues for scene geometry, and existing methods~\cite{kadambi2017depth, yoshida2018improving} have proposed physics-based approaches for depth enhancement. As such, these methods are designed for specific reflections or materials, which prevent them from being applied to general depth sensors under complex environments. Departing from existing methods, we propose a polarization-guided depth enhancement method that generalizes to diverse types of depth sensors.

Given a depth map $\mathbf{d}$ and the corresponding polarization representation $\mathbf{P}$ as described above, our method estimates the refined depth with a neural network $f$ as
\begin{equation}
    \hat{\mathbf{d}} = f(\mathbf{P}, \mathbf{d}; \theta),
\end{equation}

where $\theta$ are the parameters of the network $f$.

Following~\cite{park2020non}, we use a combination of $L_1$ and $L_2$ loss for depth supervision, shown as follows:
\begin{equation}
    \begin{split}
     \mathcal{L}(\mathbf{d^*}, \hat{\mathbf{d}})  =& \frac{1}{ |\textbf{M}_{\mathrm{depth}}|}\sum_{m \in \textbf{M}_{\mathrm{depth}}}^{}( ||\mathbf{d}^*_m-\hat{\mathbf{d}}_m||_1 \\ &+||\mathbf{d}^*_m-\hat{\mathbf{d}}_m||^2_2),
    \end{split} 
\end{equation}where $\mathbf{d}^*, \hat{\mathbf{d}}$ are the ground truth and predicted depth, respectively. $\textbf{M}_{\mathrm{depth}}$ is the mask for valid depth pixels in $\mathbf{d}^*$.

\subsection{Polarization Prompt Fusion Tuning}
\label{sec:finetuning}

\paragraph{Pre-Trained Model Weights}
Existing polarization-depth datasets are relatively small in comparison to RGB-D datasets \cite{Silberman-ECCV12, uhrig2017sparsity}. This poses a significant challenge to learning a highly generalizable model. This is more severe under our setting since there are diverse types of depth sensors, each having a unique set of measurement degradation characteristics. To overcome this, we transfer knowledge extracted from large-scale RGB-guided depth completion datasets and use it as a foundation for our model generalizability. Specifically, we do this by loading pre-trained model weights to our encoder-decoder backbone, as annotated in Figure \ref{fig:framework}. We then refine the model by incorporating polarization guidance, through a fusion path consisting of a stack of our proposed Polarization Prompt Fusion Block (PPFB). Our experiments validate that this strategy is necessary to exploit the polarization data. In this work, we use CompletionFormer~\cite{youmin2023completionformer} pre-trained on NYU-Depth V2
dataset \cite{Silberman-ECCV12} as our foundation.

\paragraph{Fusion.}

As illustrated in Figure~\ref{fig:framework}, our network first encodes each modality via three convolutional encoders. The modality-specific embeddings are then combined as a multi-modal embedding and passed as an input feature to the first Polarization Prompt Fusion Block (PPFB). Here, we treat the polarization embedding as the prompt feature. A series of PPFB modules then take each previously updated prompt as the input prompt feature, the previous encoder output as the input feature, and produce an updated feature to be passed into the next encoder layer. This operation allows the polarization information to persist and be fused in each layer of the encoder module. Lastly, the encoded features are passed through a decoder block to obtain the final enhanced depth $\mathbf{\hat d}$.

It is non-trivial to introduce another modality to a pre-trained foundation model as the polarization modality deviates drastically from the RGB one. To handle this modality misalignment, we devise a Polarization Prompt Fusion Block (PPFB). The PPFB is designed to put emphasis on the polarization input, it updates encoder features sequentially across the encoder block. Inspired by Ye \textsl{et al.}~\cite{Ye_2023_ICLR}, the proposed PPFB includes two parallel branches: (1) An input feature update branch in which the polarization prompt $\mathbf{M}$ are utilized to update and produce a fused feature $\mathbf{X^*}$ with a spatial-wise operation; (2) A polarization prompt learning branch where the input features, $\mathbf{X}$, are used to compute the updated polarization prompt $\mathbf{M^*}$ via a channel-wise fusion strategy. Inspired by the spatial fovea operation introduced by Zhu \textsl{et al.}~\cite{Zhu_2023_CVPR}, we utilize an attention-like operation for input feature enhancement. In this process, the polarization prompt and the input patch token ($C$-dimensional) are concatenated and projected by a multi-layer perceptron (MLP) to produce the key $\mathbf{k}^x$, query $\mathbf{q}^x$, and value $\mathbf{v}^x$ vectors with learnable parameters $\mathbf{W}^x_k, \mathbf{W}^x_q, \mathbf{W}^x_v \in \mathbb{R}^{2C\times C}$ as
\begin{align}
   & [\mathbf{k}^x; \mathbf{q}^x; \mathbf{v}^x] = [\mathbf{M}; \mathbf{X}] \times [\mathbf{W}^x_k; \mathbf{W}^x_q; \mathbf{W}^x_v],\\
      &  [\mathbf{X^*}; \mathbf{M^x}] = \mathrm{FC_{d}}(\lambda \cdot \mathrm{softmax}(\mathrm{FC}([\mathbf{q}^x; \mathbf{k}^x])) \odot \mathbf{v}^x),
\end{align}
where $\odot$ is the element-wise multiplication, $\mathrm{FC_d}$ is a fully-connected layer with dropout. 

With the spatial attention-like output $\mathbf{M}^x$, we learn a new polarization prompt $\mathbf{M^*}$ with a channel fovea operation, which is illustrated in Figure \ref{fig:channel-fovea}, and as follows:
\begin{align}
    &\mathbf{k}^m, \mathbf{q}^m, \mathbf{v}^m = \mathbf{X}, \mathbf{M}, \mathbf{k}^x,\\
    &[s_q; s_k] = \frac{1}{H\times W} \sum\limits_{i=1}^{H}\sum\limits_{j=1}^{W}(\mathrm{FC}(\mathbf{M}+\mathbf{X})),\\
    &\mathbf{M^*} = \mathbf{M^x}+\mathrm{FC}((s_q \mathbf{Q}^m + s_k \mathbf{K}^m)\odot \mathbf{V}^m).
\end{align}

This follows the design of Li \textsl{et al.}~\cite{Li_2019_CVPR}, in which we explore the channel-wise relationship for our prompt update by learning channel-wise statistics $s_q, s_k$ for query and key by shrinking their embedding with height and width $H, W$ through spatial dimensions. 

With PPFB, we can input our polarization modality into the pre-trained models more efficiently. We randomly initialize all the parameters of our PPFB. The entire network is then trained and updated with polarization data.

\setlength{\belowcaptionskip}{-1pt}
\begin{figure*}
  \centering
  \hspace{0.02\linewidth}
\begin{minipage}{0.1794\linewidth}
    \centering
    \footnotesize{Sensor depth}
\end{minipage}%
\begin{minipage}{0.1794\linewidth}
    \centering
    \footnotesize{GT depth}
\end{minipage}%
\begin{minipage}{0.1794\linewidth}
    \centering
    \footnotesize{DySPN~\cite{lin2022dynamic}}
\end{minipage}%
\begin{minipage}{0.1794\linewidth}
    \centering
    \footnotesize{ CompletionFormer~\cite{youmin2023completionformer}}
\end{minipage}%
\begin{minipage}{0.1794\linewidth}
    \centering
    \footnotesize{Ours}
\end{minipage}\\[1ex]
  
\begin{minipage}{0.02\linewidth}
    \rotatebox{90}{\footnotesize{Top view}}
\end{minipage}%
\begin{minipage}{0.1794\linewidth}
    \includegraphics[width=\linewidth, height=2.5cm]{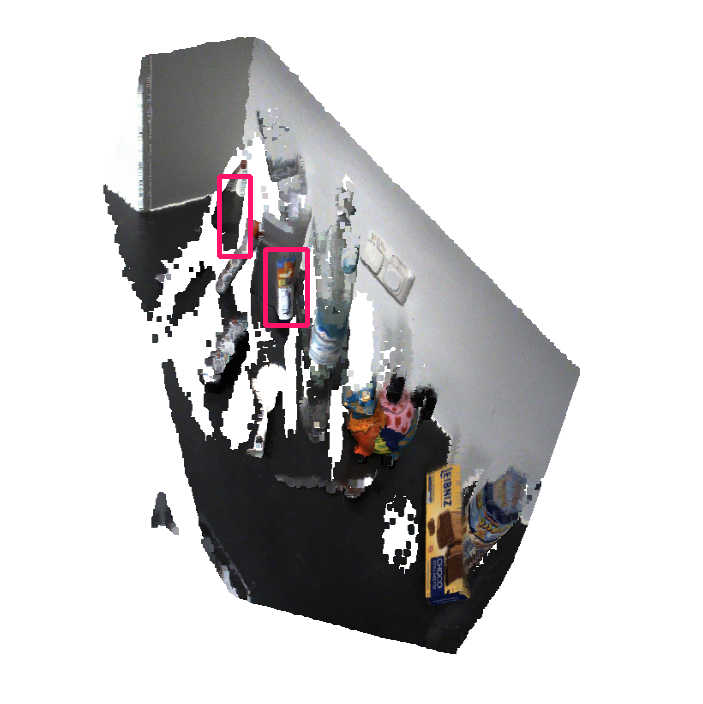}
\end{minipage}%
\begin{minipage}{0.1794\linewidth}
    \includegraphics[width=\linewidth, height=2.5cm]{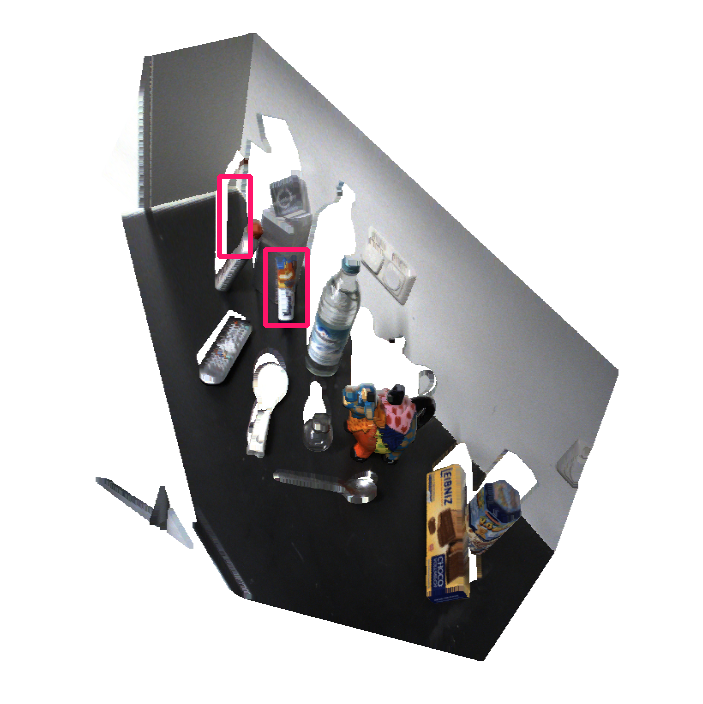}
\end{minipage}%
\begin{minipage}{0.1794\linewidth}
    \includegraphics[width=\linewidth, height=2.5cm]{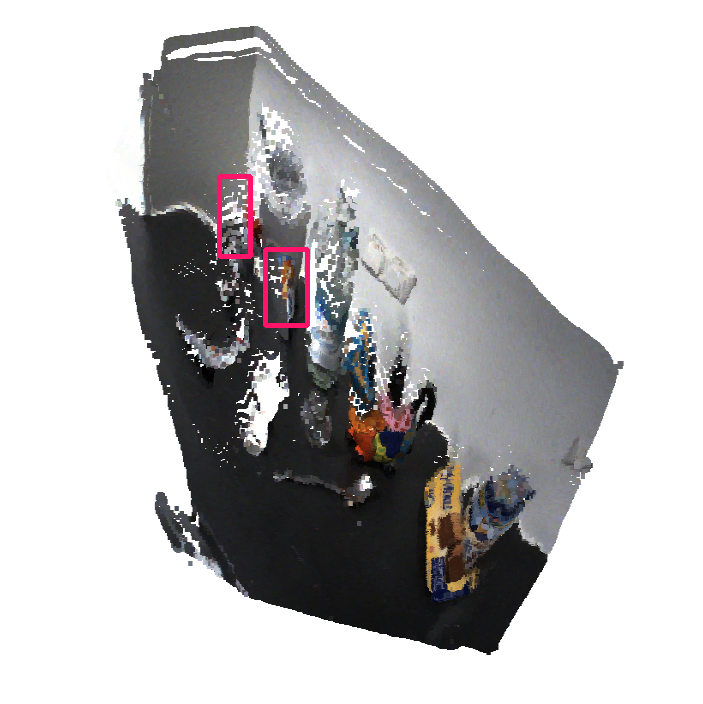}
\end{minipage}%
\begin{minipage}{0.1794\linewidth}
    \includegraphics[width=\linewidth, height=2.5cm]{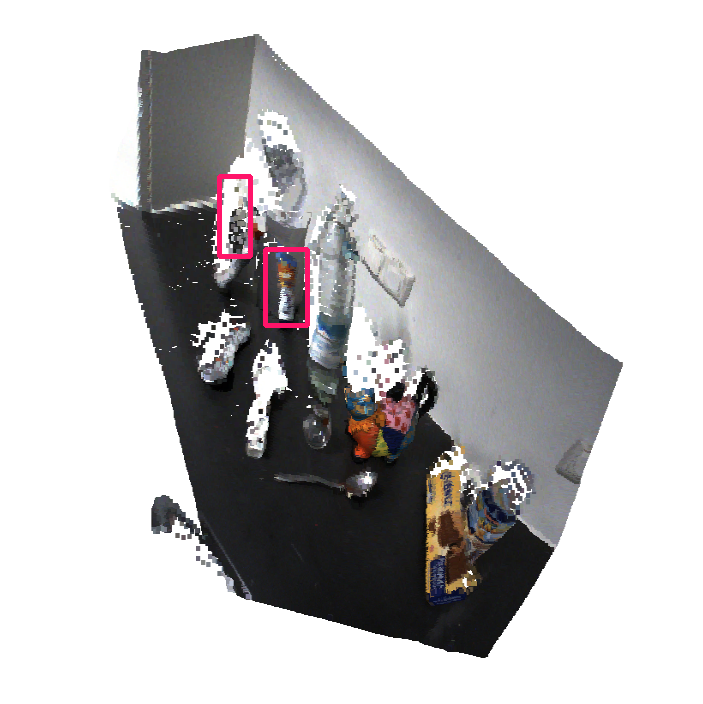}
\end{minipage}%
\begin{minipage}{0.1794\linewidth}
    \includegraphics[width=\linewidth, height=2.5cm]{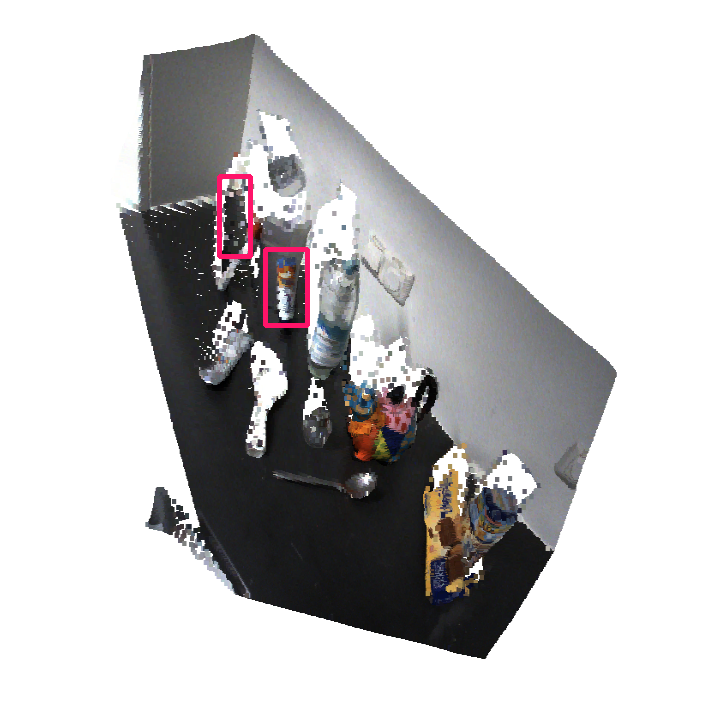}
\end{minipage}\\[1ex]

\begin{minipage}{0.02\linewidth}
    \rotatebox{90}{\footnotesize{Front view}}
\end{minipage}%
\begin{minipage}{0.1794\linewidth}
    \includegraphics[width=\linewidth]{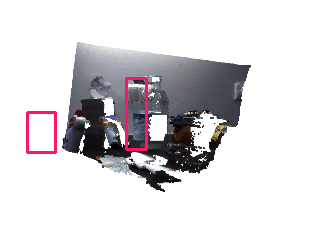}
\end{minipage}%
\begin{minipage}{0.1794\linewidth}
    \includegraphics[width=\linewidth]{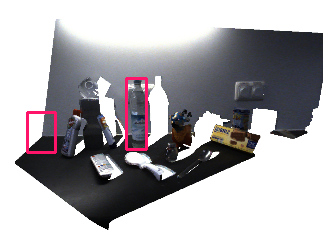}
\end{minipage}%
\begin{minipage}{0.1794\linewidth}
    \includegraphics[width=\linewidth]{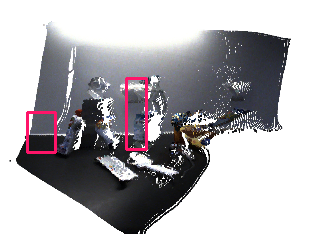}
\end{minipage}%
\begin{minipage}{0.1794\linewidth}
    \includegraphics[width=\linewidth]{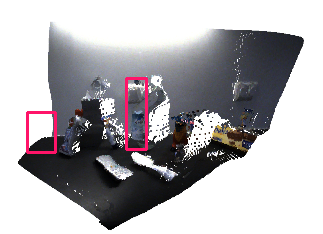}
\end{minipage}%
\begin{minipage}{0.1794\linewidth}
    \includegraphics[width=\linewidth]{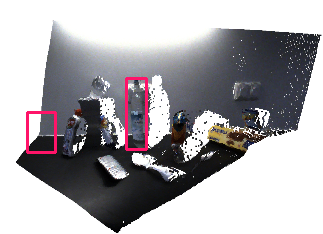}
\end{minipage}\\[1ex]
    \vspace{-2mm}
  \caption{\textbf{Comparison between our approach and baselines~\cite{lin2022dynamic,youmin2023completionformer} using point cloud visualization.} In addition to restoring challenging irregularities (e.g. the transparent bottle highlighted in row 2), we also observe a higher degree of regularity (e.g. the left-bottom corner highlighted in row 2, which is misaligned in the sensor depth, resulting in blank point clouds) using our proposed method, showing strong surface geometry information provided by polarization.}
  \label{fig:3dvis}
\end{figure*}

\setlength{\belowcaptionskip}{+1pt}
\begin{table*}[!ht]
\footnotesize
\centering
\renewcommand{\arraystretch}{1.2}
\setlength{\belowcaptionskip}{-8pt}
\resizebox{0.85\linewidth}{!}{
\begin{tabular}{ p{1.5cm} | p{4.3cm}  | >{\centering\arraybackslash}p{1.45cm} >{\centering\arraybackslash}p{1.4cm} >{\centering\arraybackslash}p{1.4cm} >{\centering\arraybackslash}p{1.4cm} >{\centering\arraybackslash}p{1.4cm}}
\toprule[1pt]

\multirow{2}{*}{Sensor Type} & \multirow{2}{*}{Model} & RMSE (mm)$\downarrow$ & MAE (mm)$\downarrow$  & $\delta_1 \uparrow$ &$\delta_2 \uparrow $ & $\delta_3 \uparrow $\\
\hline

\multirow{6}{*}{Active Stereo} & Polarized 3D* \cite{kadambi2015polarized}  & 755.64 & 659.83 & 0.077 & 0.164 & 0.284 \\
& SemAttNet \cite{nazir2022semattnet} & 56.36 & 36.68 & 0.975 & 0.998 & 0.999 \\
& DySPN \cite{lin2022dynamic} & 54.56 & 32.22 & 0.975 & 0.997 & 0.998 \\

 & CompletionFormer \cite{youmin2023completionformer} & 32.78 & 15.64 & 0.991 & 0.999 & 0.999    \\

 & $\text{CompletionFormer}^\dag$~\cite{youmin2023completionformer} & 33.86  & 15.68 & 0.991 & 0.991 & 0.999   \\

 & Ours & \textbf{24.78} & \textbf{10.66} & \textbf{0.995} & \textbf{0.999} & \textbf{1.000}\\
\hline

\multirow{6}{*}{d-ToF} & Polarized 3D* \cite{kadambi2015polarized}  & 851.86 & 777.43 & 0.034 & 0.073 & 0.130 \\
& SemAttNet \cite{nazir2022semattnet} & 54.32 & 33.92 & 0.977 & 0.999 & 0.999 \\
& DySPN \cite{lin2022dynamic} & 51.82 & 31.19 & 0.977 & 0.997 & 0.998  \\

 & CompletionFormer \cite{youmin2023completionformer} & 33.93 & 15.10 & 0.987 & 0.999 & 0.999  \\
 & $\text{CompletionFormer}^\dag$~\cite{youmin2023completionformer} & 36.40 & 17.22 & 0.984 & 0.991 & 0.999   \\

  & Ours & \textbf{23.84} & \textbf{10.47} & \textbf{0.995} & \textbf{0.999} & \textbf{1.000}\\

\hline

\multirow{6}{*}{i-ToF} & Polarized 3D* \cite{kadambi2015polarized}  & 709.97 & 629.98 & 0.035 & 0.074 & 0.123 \\
& SemAttNet \cite{nazir2022semattnet} & 81.35 & 56.49 & 0.934 & 0.997 & 0.999 \\
& DySPN \cite{lin2022dynamic} & 69.94 & 48.11 & 0.949 & 0.997 & 0.999\\

 & CompletionFormer \cite{youmin2023completionformer} & 49.95 & 29.08 & 0.974 & 0.998 & 0.999  \\
  & $\text{CompletionFormer}^\dag$~\cite{youmin2023completionformer} & 54.94 & 32.75 & 0.971 & 0.998 & 0.999  \\

    & Ours & \textbf{37.27} & \textbf{21.37} & \textbf{0.987} & \textbf{0.999} & \textbf{1.000}\\

  \hline
  
  \multirow{6}{*}{All} & Polarized 3D* \cite{kadambi2015polarized}  & 772.49 & 689.08 & 0.048 & 0.103 & 0.179 \\
& SemAttNet \cite{nazir2022semattnet} & 64.01 & 42.36 & 0.962 & 0.998 & 0.999 \\
& DySPN \cite{lin2022dynamic} & 58.77 & 37.17 & 0.967 & 0.997 & 0.998\\

 & CompletionFormer \cite{youmin2023completionformer} & 38.88 & 19.94 & 0.984 & 0.998 & 0.999  \\
  & $\text{CompletionFormer}^\dag$~\cite{youmin2023completionformer} & 41.73 & 21.88 & 0.982 & 0.993 & 0.999  \\

  & Ours & \textbf{28.63} & \textbf{14.17} & \textbf{0.992} & \textbf{0.999} & \textbf{1.000}\\

\toprule[1pt]
\end{tabular}}
  \vspace{-2mm}
\caption{\textbf{Quantitative comparison on the HAMMER~\cite{hammer} dataset.} To facilitate fair comparisons, we pass polarization data to the RGB-guided models~\cite{nazir2022semattnet,lin2022dynamic,youmin2023completionformer}. Sensor depths are obtained from an active stereo camera, d-ToF sensor, and i-ToF sensor. Our method achieves the best performance consistently in all cases. * indicates that directly applying Polarized 3D~\cite{kadambi2015polarized} with author provided code fails in our scenario. $\dag$ denotes the model is fine-tuned with RGB images. We train on all data and test on different sensor types for analysis.}

\label{table:performance}
\end{table*}
\setlength{\belowcaptionskip}{+8pt}

\setlength{\belowcaptionskip}{-5pt}

\begin{figure*}
    \footnotesize
    \centering
    \begin{minipage}{0.005\linewidth}
        \rotatebox{90}{Stereo \hspace{1.4cm} d-ToF \hspace{1.2cm}
      i-ToF}
      \end{minipage}%
      \hspace{0.01\linewidth}
      \begin{minipage}{0.98\linewidth}
    \includegraphics{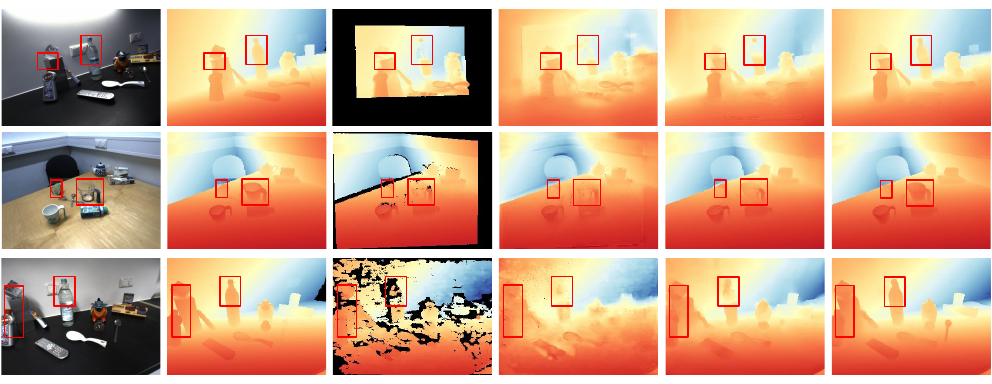}
      \end{minipage}\\[.5ex]
      \begin{minipage}{0.9\linewidth}
      \, RGB image \hspace{1.0cm} Ground truth depth \hspace{0.8cm} Sensor depth \hspace{1.3cm} DySPN \cite{lin2022dynamic} \hspace{0.65cm} CompletionFormer \cite{youmin2023completionformer} \hspace{1cm} Ours
  \end{minipage}%
    \caption{\textbf{Qualitative comparison on HAMMER dataset~\cite{hammer}}. We present results on different depth sensors. The red boxes highlight regions to emphasize. We can observe significantly better results on transparent surfaces, i.e. the water bottle, and detailed regions, for example, the stack of objects on the left in Row 2.}
    \label{fig:comp_baseline}
\end{figure*}
\setlength{\belowcaptionskip}{+5pt}

\section{Experiments}
\label{sec:experiments}

\subsection{Experimental Setup}
\paragraph{Dataset} We evaluate our method on the public real-world HAMMER dataset~\cite{hammer}. The HAMMER dataset is a polarization-depth dataset consisting of 13 different scenes, with 7200 distinct frames capturing scenes with objects in total. It aids our analysis as it contains both depths from commodity-level depth sensors and paired hyper-quality rendered ground truth depths of the captured scene. Specifically, the dataset includes 3 types of depth sensor measurement: stereo camera depths (D435), d-ToF sensor (L515), and i-ToF sensor depths. This variety of real-world depths allows us to study different depth sensor degradation. We use the official training set for training, which consists of the first nine scenes, giving 5253 frames in total. Further, we split the remaining data into two scenes for validation and two scenes for testing.

\vspace{-6pt}
\paragraph{Baselines} There is no existing polarization-based depth enhancement approach that is tackling the proposed setup. Specifically, given polarization guidance, the task is to enhance commodity-level depth sensor measurement capturing real-world scenes, regardless of the sensing technology used (active stereo, d-ToF, i-ToF, etc.). Thus, we first compare against Polarized 3D~\cite{kadambi2015polarized}, an object-level physics-based depth enhancement method. In addition, we adopt 3 state-of-the-art RGB-based depth completion methods as baselines. Namely, we compare with CompletionFormer~\cite{youmin2023completionformer}, SemAttNet~\cite{nazir2022semattnet}, and DySPN~\cite{lin2022dynamic}, which are learning-based approaches that receive the RGB images as input guidance. We change the network input from RGB images to polarization images and train the models on the same dataset for fair comparison. The fairness is in the sense that polarization information is given to all models. Note that SemAttNet~\cite{nazir2022semattnet} requires additional instance segmentation maps, which are provided in the Hammer dataset.
\raggedbottom

\subsection{Assessment}

Table \ref{table:performance} presents the quantitative results of all methods evaluated on the HAMMER~\cite{hammer} dataset. We evaluate our method on each depth sensor type separately to validate the generalizability across various kinds of degradation. Our approach achieves the best performance consistently on all sensor depths and on all metrics. Specifically, compared to CompletionFormer~\cite{youmin2023completionformer}, the proposed method obtains a performance gain of 24.4\%, 29.7\%, 25.3\% on the RMSE metric; and 31.8\%, 30.6\%, 26.5\% on the MAE metric, evaluated on stereo, d-ToF, and i-ToF benchmarks respectively. The prediction results on i-ToF sensors are relatively worse compared to those on active stereo and d-ToF, this is due to the large area of missing depths that require in-painting since i-ToF sensors have smaller field-of-view (FoV) compared to other devices.

Figure \ref{fig:comp_baseline} reports the qualitative comparisons of the best three baselines. The proposed method improves the quality of depth maps from different depth sensors. The results of DySPN~\cite{lin2022dynamic} are blurry, especially at regions with raw, shape-incorrect depth measurements. Further, the enhanced depth lacks accuracy at detailed regions, this is more apparent in the stereo depth maps. While CompletionFormer~\cite{youmin2023completionformer} produces poor boundaries, our enhanced depths are more robust to irregularities that are hard to generalize using RGB guidance, given a small training dataset. Further, the proposed method is able to correct false surface depth in abnormal conditions, such as transparent object surfaces. As the point cloud results in Figure \ref{fig:3dvis} present, our method generates highly regularized depth, for example, the flat walls. Hence we argue that our method can learn the underlying surface normal information from polarization cues. Further, as the stereo (row 1) and d-ToF (row 2) results in Figure \ref{fig:comp_baseline} illustrate, our method separates foreground and background surfaces adequately (e.g., the teapot and the wall). In the i-ToF case (row 3), our method presents the ability to generalize to the unseen surface, obtaining a smoother transition between the known and unknown regions.

\subsection{Analysis} \label{ablations}
In the following, we conduct ablation experiments on the average of all sensor types to evaluate each component in our method. We adopt the same setting for all experiments and the details are included in the supplementary.

\vspace{-7pt}

\begin{figure}[!t]
  \centering
  \includegraphics[width=.7\linewidth]{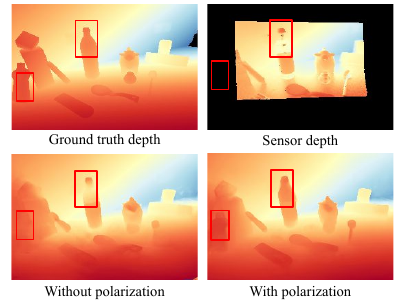}
  \vspace{-2mm}
  \caption{\textbf{Qualitative results for the ablation study on polarization input.} We can observe a significant performance drop given no polarization information, especially in detailed regions.}
  \label{fig:Ablation study B}
\end{figure}

\paragraph{Polarization Guidance} We first analyze the importance of polarization information to depth enhancement. In Table~\ref{table:ablation_input}, we demonstrate the results using our proposed architecture, with and without polarization input during both training and testing. Note that we only substitute the polarization data by the RGB image at the input layer, without modifying the rest of the proposed model structure. We observe that having no polarization guidance results in inferior performance. Using polarization cues improves the performance substantially, validating the proposed method. Figure~\ref{fig:Ablation study B} further illustrates that, with polarization as guidance, the proposed method achieves accurate estimation of the transparent regions. 

\vspace{-7pt}

\begin{table}[t]
\footnotesize
\centering
\renewcommand{\arraystretch}{1.2}
\setlength{\belowcaptionskip}{-2pt}
\resizebox{0.85\linewidth}{!}{
\begin{tabular}{p{2.4cm}   | >{\centering\arraybackslash}p{0.6cm} >{\centering\arraybackslash}p{0.6cm} >{\centering\arraybackslash}p{0.6cm} >{\centering\arraybackslash}p{0.6cm} >{\centering\arraybackslash}p{0.6cm}}
\toprule[1pt]
 PPFT Guidance &  RMSE$\downarrow$ & MAE$\downarrow$  & $\delta_1 \uparrow$ &$\delta_2 \uparrow $ & $\delta_3 \uparrow $\\
\hline
    RGB &    33.06 & 16.55 & 0.989 & 0.999 & 0.999\\
  Polarization &  \textbf{28.63} & \textbf{14.17}	& \textbf{0.992} & \textbf{0.999} & \textbf{1.000} \\
\toprule[1pt]
\end{tabular}}
  \vspace{-2mm}

\caption{\textbf{Ablation experiment on the polarization guidance.} By using RGB guidance in PPFT, the performance drops significantly, showing the importance of polarization as geometry guidance.}
\label{table:ablation_input}
\end{table}

\setlength{\belowcaptionskip}{+6pt}

\begin{table}[t]
\footnotesize
\centering
\renewcommand{\arraystretch}{1.2}
\setlength{\belowcaptionskip}{-13pt}
\resizebox{0.85\linewidth}{!}{

\begin{tabular}{ p{2.9cm} | p{0.77cm} p{0.77cm} p{0.77cm} p{0.77cm}}
\toprule[1pt]

 \multirow{2}{*}{Method} & \multicolumn{1}{c}{RMSE} & \multicolumn{1}{c}{$\delta$ RMSE} & \multicolumn{1}{c}{MAE} & \multicolumn{1}{c}{$\delta$ MAE} \\
  & \multicolumn{1}{c}{(mm)$\downarrow$} & \multicolumn{1}{c}{(mm)$\downarrow$} & \multicolumn{1}{c}{(mm)$\downarrow$} & \multicolumn{1}{c}{(mm)$\downarrow$} \\
 
\hline
 CompletionFormer*~\cite{youmin2023completionformer} & 313.60  & - & 241.40 & - \\
\hline
 
 Without PPFT & 41.17  & - & 22.14 & - \\
 
 Ours & \textbf{28.63} & \textbf{-12.54} & \textbf{14.17}	& \textbf{-7.97}  \\
 
\toprule[1pt]
\end{tabular}}

  \vspace{-2mm}

\caption{\textbf{Ablation experiments for cross-modal transfer learning strategy.} With our proposed Polarization Prompt Fusion Tuning (PPFT) the performance is improved significantly. * denotes that the foundation model is pre-trained on NYU-Depth V2 dataset~\cite{Silberman-ECCV12}. $\delta$ here indicates the performance gain of our method compared to the case \textbf{Without PPFT}.}
\label{table:cross_modality}
\end{table}

\setlength{\belowcaptionskip}{+13pt}

\setlength{\belowcaptionskip}{-1pt}
\setlength{\belowcaptionskip}{+1pt}

\paragraph{Cross-modal Transfer Learning} Our cross-modal transfer learning design is the key to achieving superior performance on real data. The existing polarization-based depth dataset is significantly smaller than RGB-based datasets, preventing most data-driven methods from achieving satisfactory performance. As Table \ref{table:cross_modality} reports, the proposed cross-modal transfer learning module improves the performance significantly compared with training the model on the HAMMER\cite{hammer} dataset from scratch. Without the proposed Polarization Prompt Fusion Tuning (PPFT) strategy, the model is not able to fit such a complex dense geometry prediction problem. In this regard, our proposed method is able to effectively utilize the limited polarization information, adapting well to sensor depth in complex scenes.

\begin{table}[t]
\footnotesize
\centering
\renewcommand{\arraystretch}{1.2}
\setlength{\belowcaptionskip}{-14pt}

\begin{tabular}{p{2.5cm} | >{\centering\arraybackslash}p{0.6cm} >{\centering\arraybackslash}p{0.6cm} >{\centering\arraybackslash}p{0.6cm} >{\centering\arraybackslash}p{0.6cm} >{\centering\arraybackslash}p{0.6cm}}
\toprule[1pt]
 Fusion Method & RMSE$\downarrow$ & MAE$\downarrow$  & $\delta_1 \uparrow$ &$\delta_2 \uparrow $ & $\delta_3 \uparrow $\\
\hline
 
  Early Fusion~\cite{feng2020deep} & 30.33  & 15.30 & 0.991 & 0.999 & 1.000 \\

Ours (PPFB-shallow) & 29.77 & 14.79	& 0.991 & 0.999 & 0.999 \\

  Ours (PPFB-deep) & \textbf{28.63} & \textbf{14.17}	& \textbf{0.992} & \textbf{0.999} & \textbf{1.000} \\
\toprule[1pt]
\end{tabular}
  \vspace{-2mm}
\caption{\textbf{Quantitative comparison on different fusion strategies.} It is vital to inject the PPFB module into different layers of the neural network instead of only the first layer. }
\label{table:ablation_fusion}
\end{table}

\setlength{\belowcaptionskip}{+14pt}

\paragraph{Polarization Fusion Module} We evaluate our Polarization Prompt Fusion Block (PPFB) with respect to multi-modal data fusion. We compare an early fusion strategy~\cite{feng2020deep} with naive polarization concatenation and two variants of our proposed method, (a) a shallow (only a single PPFB at the first encoder layer), and (b) a deep version (i.e. our full model). Quantitative results are presented in Table \ref{table:ablation_fusion}. This result illustrates that with our full-depth parallel branches design, the additional modality can be forwarded into deep layers of the pre-trained foundation more effectively, enabling a thorough fusion with the input features.

\begin{figure}[t]
\centering
\resizebox{0.85\linewidth}{!}{
\begin{tabular}{@{}c@{\hspace{1.4mm}}c@{\hspace{1.4mm}}c@{\hspace{1.4mm}}c@{}}

\includegraphics[width=0.3\linewidth]{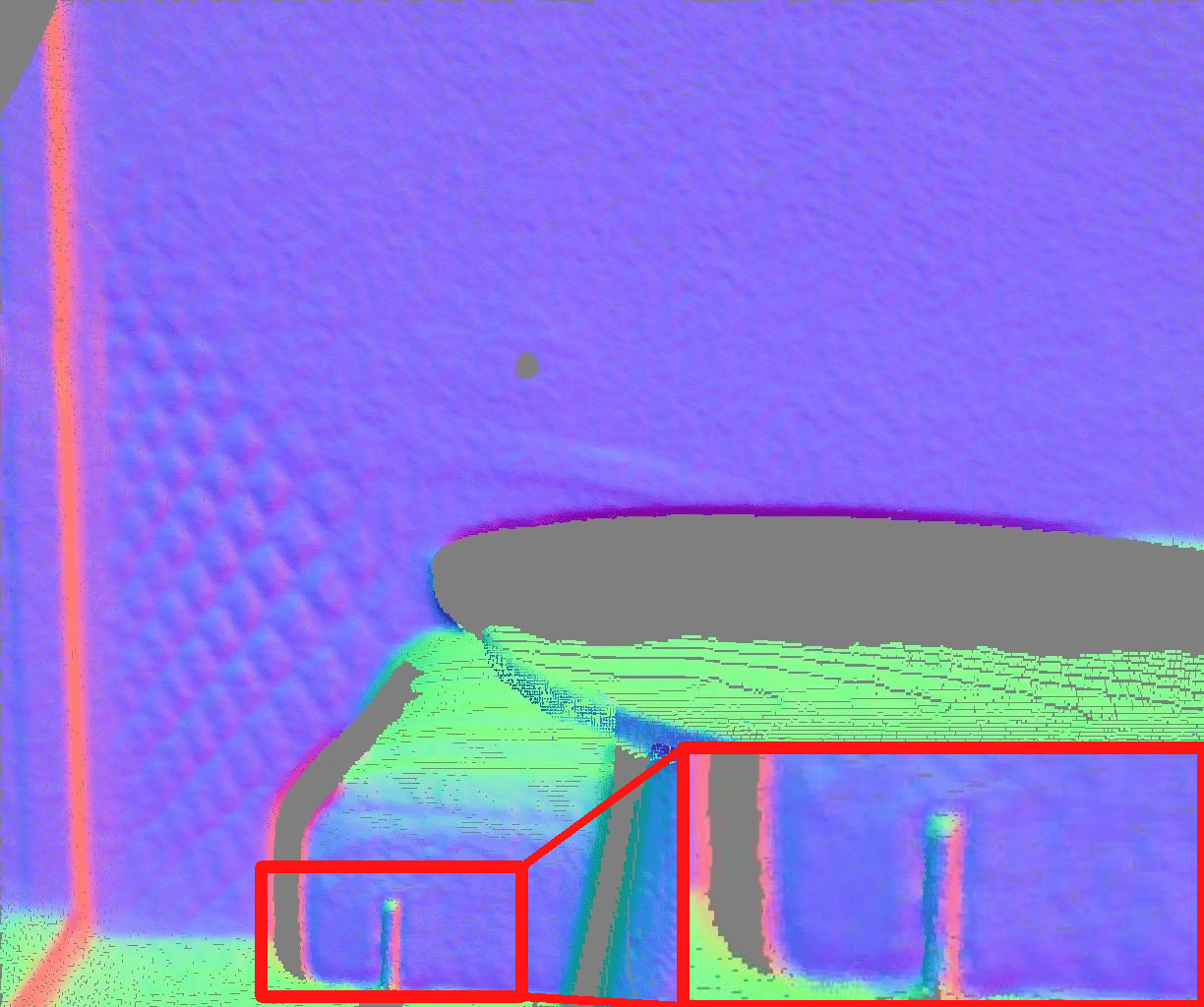}&
\includegraphics[width=0.3\linewidth]{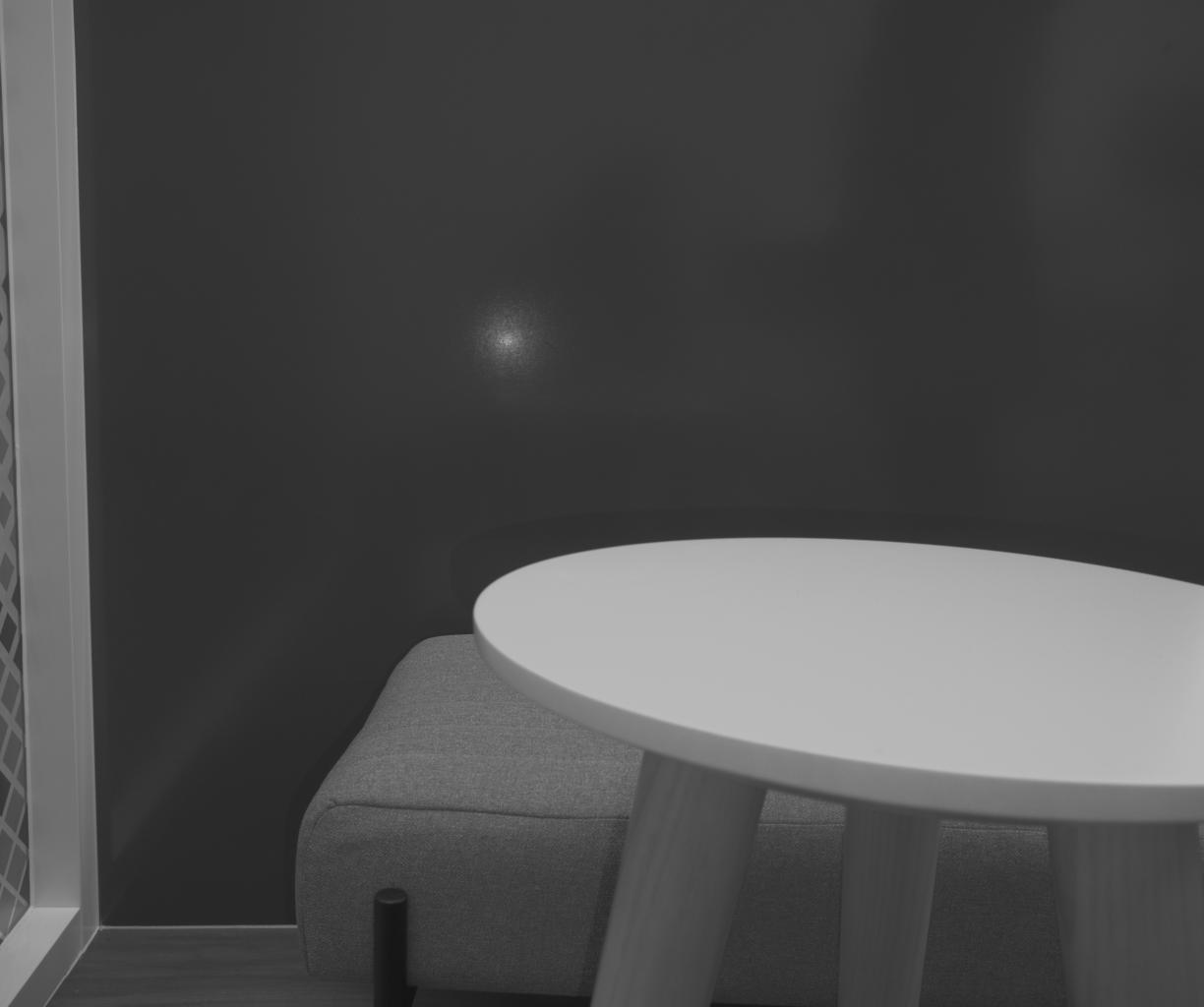}& \includegraphics[width=0.3\linewidth]{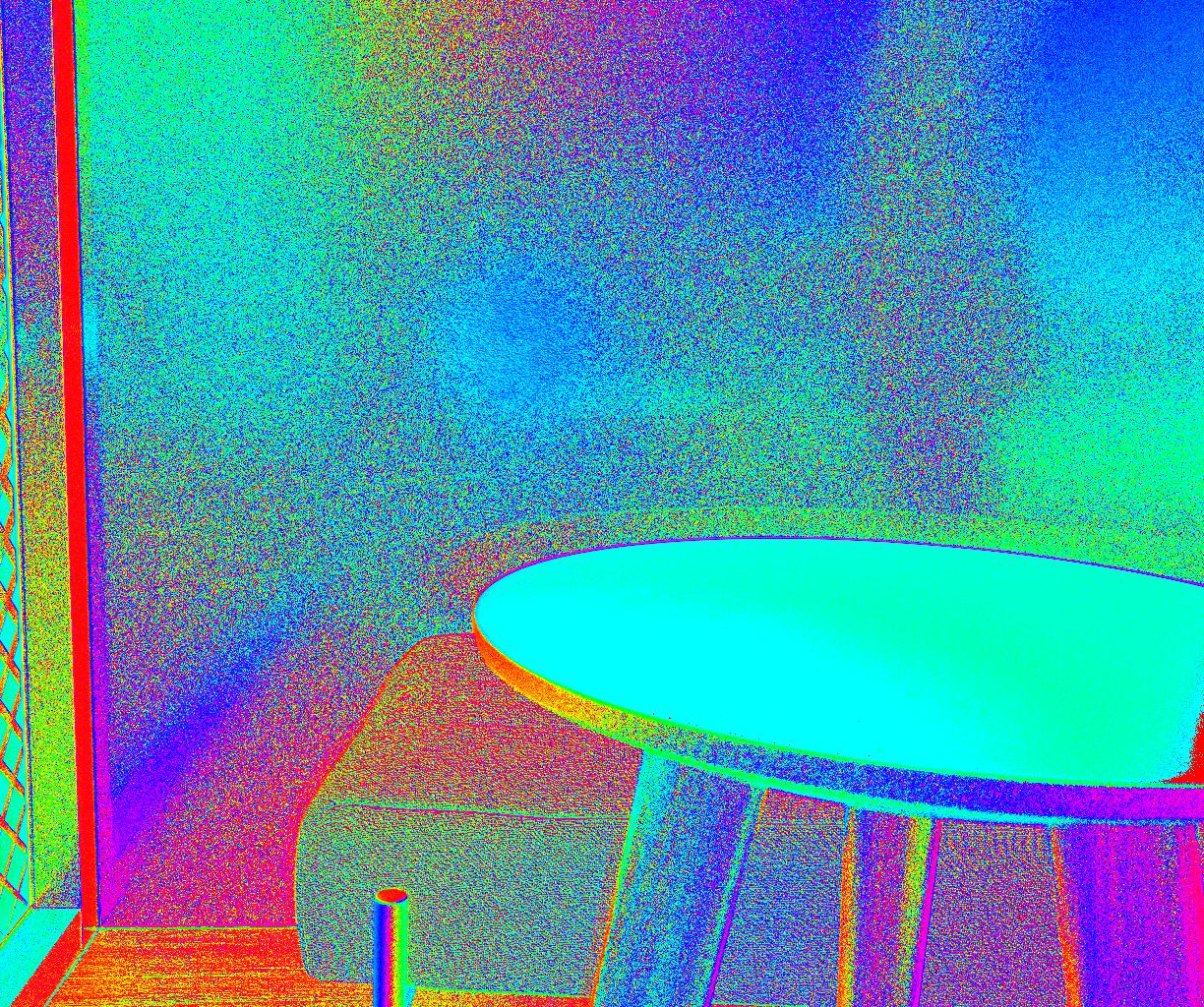}\\

  \normalsize{GT normal} &  \normalsize{Intensity} &\normalsize{AoLP} & \\ 
   
 \includegraphics[width=0.3\linewidth]{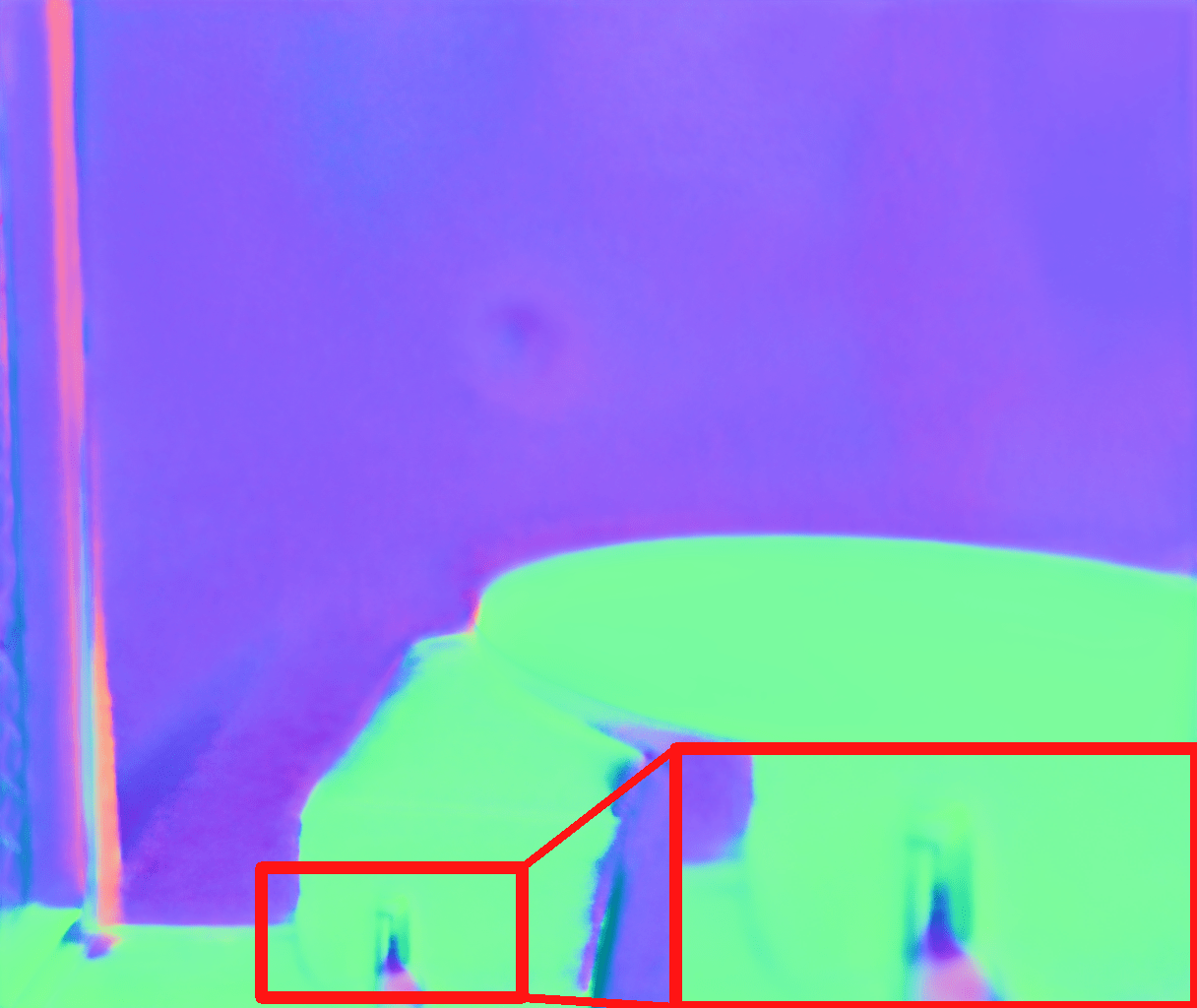}&
\includegraphics[width=0.3\linewidth]{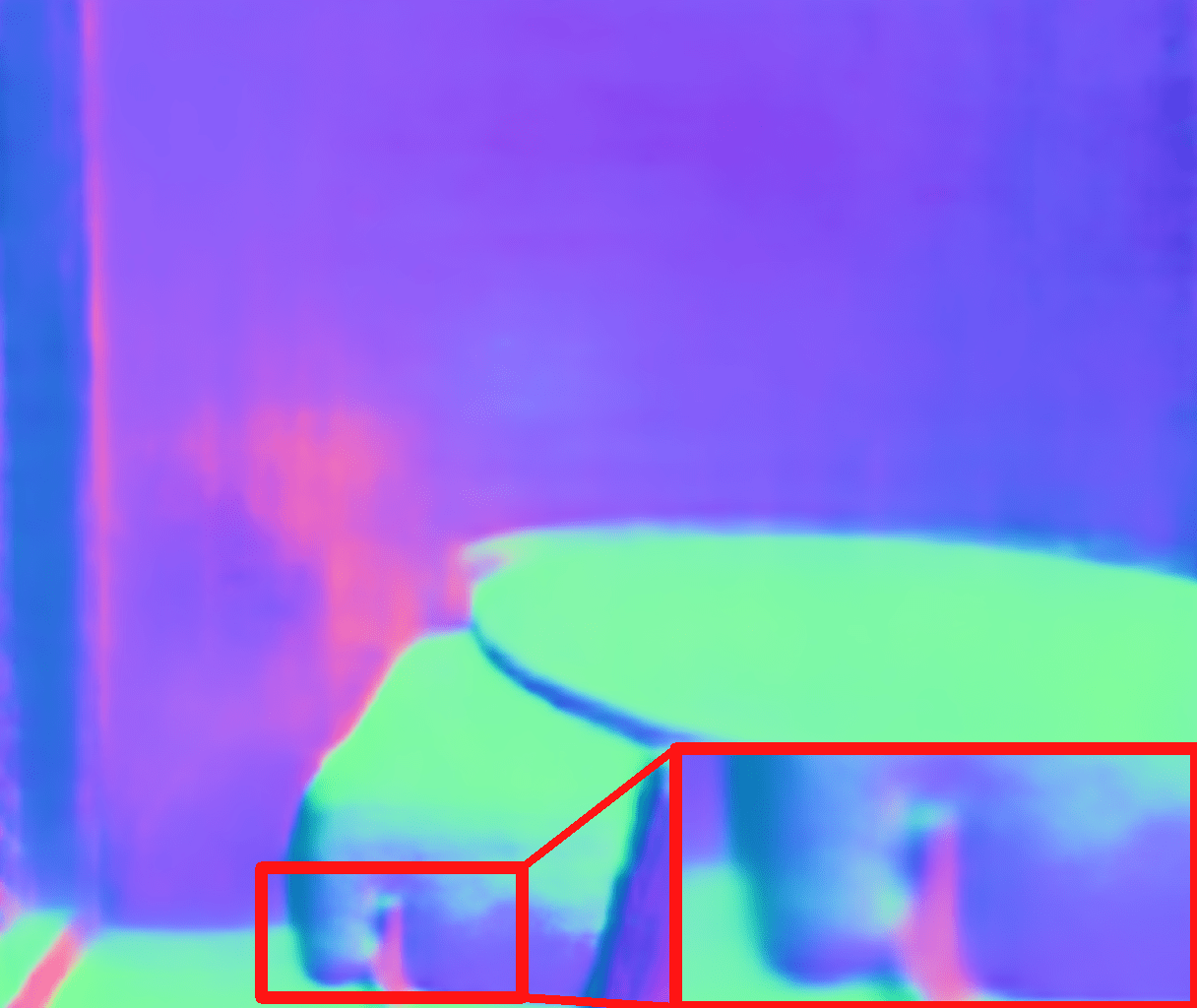}& \includegraphics[width=0.3\linewidth]{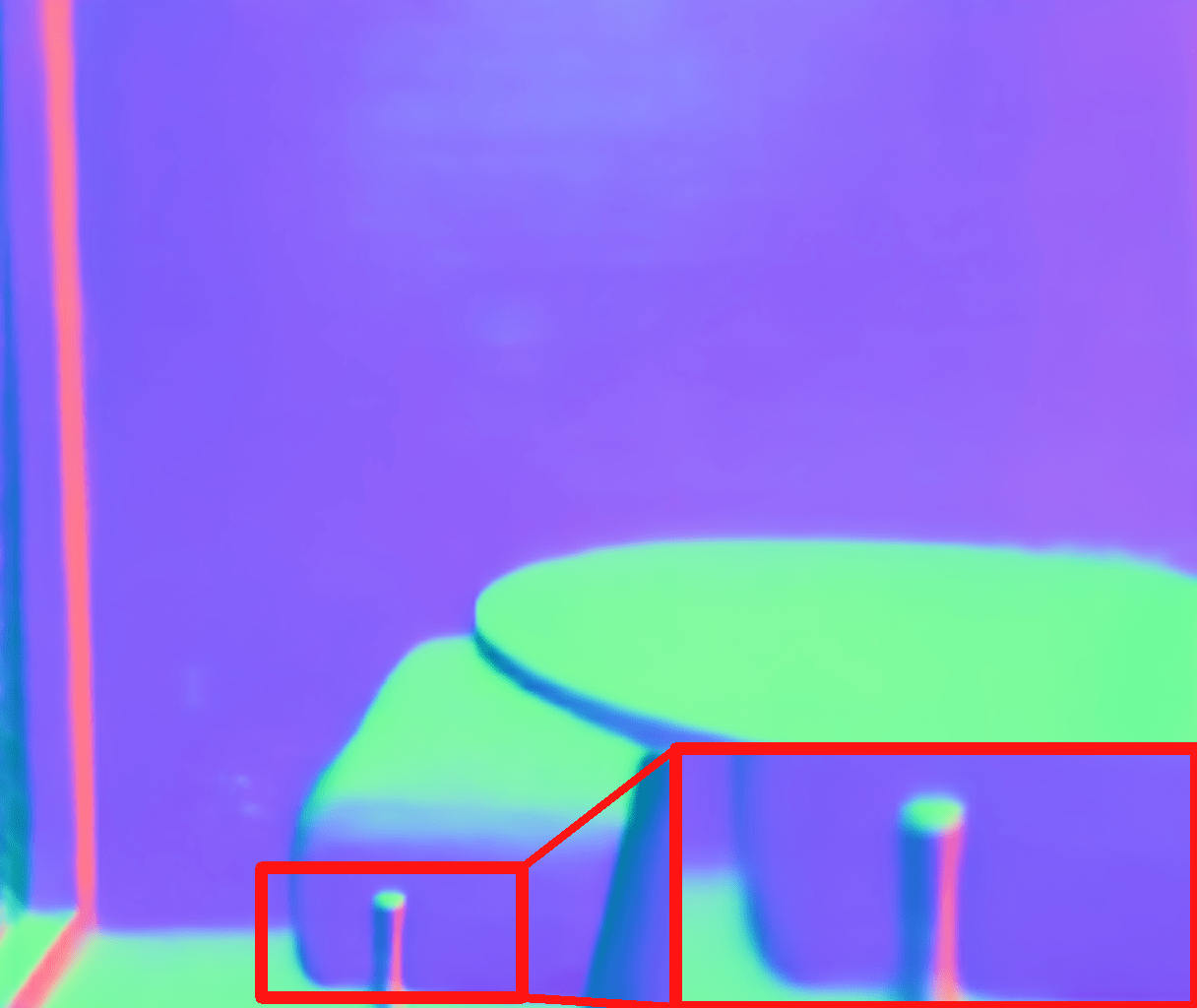}&\\

 \normalsize{SfP-Wild \cite{lei2022shape}} &  \normalsize{iDisc \cite{piccinelli2023idisc}} & \normalsize{iDisc \cite{piccinelli2023idisc} + PPFT}
 \end{tabular}}
\\
\vspace{-2mm}
  \caption{\textbf{Qualitative results of shape-from-polarization methods on the SPW dataset \cite{lei2022shape}}. Our proposed PPFT can be combined with other polarization-based tasks to achieve better performance on limited polarization training data. Here SfP-Wild~\cite{lei2022shape} is the state-of-the-art shape from the polarization approach. We use iDisc~\cite{piccinelli2023idisc} as the backbone with the same polarization input and achieve better performance. In this example, our proposed PPFT with iDisc~\cite{piccinelli2023idisc} produces significantly cleaner results, where the details are clearly visible as shown in the highlight region.}
\label{fig:normal_exp}
\vspace{-3mm}
\end{figure}

\subsection{PPFT for Surface Normal Estimation} We highlight the versatility of the proposed PPFT strategy to a wider range of polarization vision tasks, for example, Shape-from-Polarization (SfP) \cite{ba2020deep, lei2022shape}. As Table~\ref{table:normal_exp} shows our method can further improve the performance of the SfP method by incorporating our proposed PPFT strategy and a pre-trained RGB-based normal estimation model. Specifically, we apply our method to iDisc \cite{piccinelli2023idisc}, a foundation model trained on NYU-Depth V2 dataset~\cite{Silberman-ECCV12} that consists of 30k training samples. We then train and evaluate on SPW dataset \cite{lei2022shape}, a scene-level SfP dataset, which in comparison involves 400 training frames. Our method produces high-accuracy surface normals with fine details as shown in Figure~\ref{fig:normal_exp}, emphasizing the potential applications of the proposed PPFT module to more polarization vision tasks.

\setlength{\belowcaptionskip}{-2pt}
\begin{table}[ht]
\footnotesize
    \centering

    \vspace{-5pt}
    \renewcommand{\arraystretch}{1.1}
    \resizebox{0.85\linewidth}{!}{

    \begin{tabular}{p{2.05cm} | >{\centering\arraybackslash}p{0.54cm} >{\centering\arraybackslash}p{0.54cm} >{\centering\arraybackslash}p{0.54cm} >{\centering\arraybackslash}p{0.54cm} >{\centering\arraybackslash}p{0.54cm} >{\centering\arraybackslash}p{0.54cm}}
        \toprule
        \multirow{2}{*}{Model} & Mean & Med & RMSE & $11.5^{\circ}$ & $22.5^{\circ}$ & $30^{\circ}$ \\
         & \multicolumn{3}{c}{Angular Error $\downarrow$} & \multicolumn{3}{c}{Percentage\% $\uparrow$} \\
        \hline

         iDisc* \cite{piccinelli2023idisc} & 43.56 & 38.46 &	51.92 & 16.61 & 35.48 & 43.56 \\

        \hline

        SfP-Wild~\cite{lei2022shape} & 18.59 & 14.46 & 24.08 & 45.38 & 75.86 & 83.81\\
        
        iDisc~\cite{piccinelli2023idisc} & 12.99 & 10.11 & 16.85 & 59.68 & 87.78 & 92.48 \\
        iDisc~\cite{piccinelli2023idisc} + PPFT &  \textbf{10.91} &	\textbf{8.92} &	\textbf{14.03} &	\textbf{67.82}	& \textbf{92.12} &	\textbf{95.77} \\
        \bottomrule
    \end{tabular}}
      \vspace{-2mm}
    \caption{\textbf{Quantitative results on shape-from-polarization.} We achieve the best surface normal prediction performance by applying our PPFT strategy on iDisc \cite{piccinelli2023idisc} pre-trained on NYU Depth V2 \cite{Silberman-ECCV12}. For comparison, we pass polarization data to train iDisc with naive concatenation. *: this model is the RGB-based foundation model that is pre-trained on NYU-Depth V2 dataset~\cite{Silberman-ECCV12}.}
    \label{table:normal_exp}
    \vspace{-12pt}
\end{table}

\setlength{\belowcaptionskip}{+2pt}
\section{Conclusion}
\label{sec:conclusion}

In this work, we introduce a unified learning-based method that utilizes polarization information to improve sensor depth quality. To target the data scarcity problem with the existing polarization-depth dataset, we propose a Polarization Prompt Fusion Tuning (PPFT) strategy, which utilizes models pre-trained on large-scale RGB-D datasets. This allows both efficient use of pre-trained weights and effective fusion of polarization data. With PPFT, we achieve generalization to various depth-sensing approaches. Experiments validate substantial depth quality improvements for diverse depth inputs. Furthermore, we showcase the capability of applying the proposed method to other polarization-based vision tasks. As such, we hope that our work can contribute to the advancement of polarization-based depth enhancement and its potential applications in autonomous driving, 3D reconstruction, and beyond.

\paragraph{Acknowledgements} This work was supported by the InnoHK program.

\small
\bibliographystyle{ieeenat_fullname}
\bibliography{main.bib}

\end{document}